\documentclass[10pt,twocolumn,letterpaper]{article}

\usepackage{iccv}              
\usepackage{multirow}
\usepackage{colortbl}
\usepackage{makecell}
\usepackage{booktabs}
\usepackage{amssymb}
\usepackage{amsmath}
\usepackage[ruled,linesnumbered]{algorithm2e}
\usepackage{pifont} 
\usepackage{xcolor}

\definecolor{iccvblue}{rgb}{0.21,0.49,0.74}
\usepackage[pagebackref,breaklinks,colorlinks,allcolors=iccvblue]{hyperref}

\def\paperID{12437} 
\def\confName{ICCV}
\def\confYear{2025}

\title{Hyper3D: Efficient 3D Representation via Hybrid Triplane and  Octree Feature \\ for Enhanced 3D Shape Variational Auto-Encoders}

\author{Jingyu Guo$^{1,2}$,\;
Sensen Gao$^{3}$,\;
Jia-Wang Bian$^{3}$,\;
Wanhu Sun$^{2,4}$,\;\\
Heliang Zheng$^{2}$,\;
Rongfei Jia$^{2}$,\;
Mingming Gong$^{1,3}$\\
\normalsize{$^{1}$ The University of Melbourne \quad $^{2}$ Sensory Universe \quad $^{3}$ Mohammed Bin Zayed University of Artificial Intelligence} \\
~~\normalsize{$^{4}$ The Chinese University of Hong Kong, ShenZhen}\\
}

\begin{document}
\maketitle

\begin{abstract}
Recent 3D content generation pipelines often leverage Variational Autoencoders (VAEs) to encode shapes into compact latent representations, facilitating diffusion-based generation.
Efficiently compressing 3D shapes while preserving intricate geometric details remains a key challenge.
Existing 3D shape VAEs often employ uniform point sampling and 1D/2D latent representations, such as vector sets or triplanes, leading to significant geometric detail loss due to inadequate surface coverage and the absence of explicit 3D representations in the latent space.
Although recent work explores 3D latent representations, their large scale hinders high-resolution encoding and efficient training.
Given these challenges, we introduce Hyper3D, which enhances VAE reconstruction through efficient 3D representation that integrates hybrid triplane and octree features.
First, we adopt an octree-based feature representation to embed mesh information into the network, mitigating the limitations of uniform point sampling in capturing geometric distributions along the mesh surface.
Furthermore, we propose a hybrid latent space representation that integrates a high-resolution triplane with a low-resolution 3D grid. This design not only compensates for the lack of explicit 3D representations but also leverages a triplane to preserve high-resolution details.
Experimental results demonstrate that Hyper3D outperforms traditional representations by reconstructing 3D shapes with higher fidelity and finer details, making it well-suited for 3D generation pipelines.
Code is available at \href{https://github.com/jingyu198/Hyper3D}{https://github.com/jingyu198/Hyper3D}.
\end{abstract}
\section{Introduction}
\begin{figure}[t]
    \centering
    \includegraphics[width=0.95\linewidth]{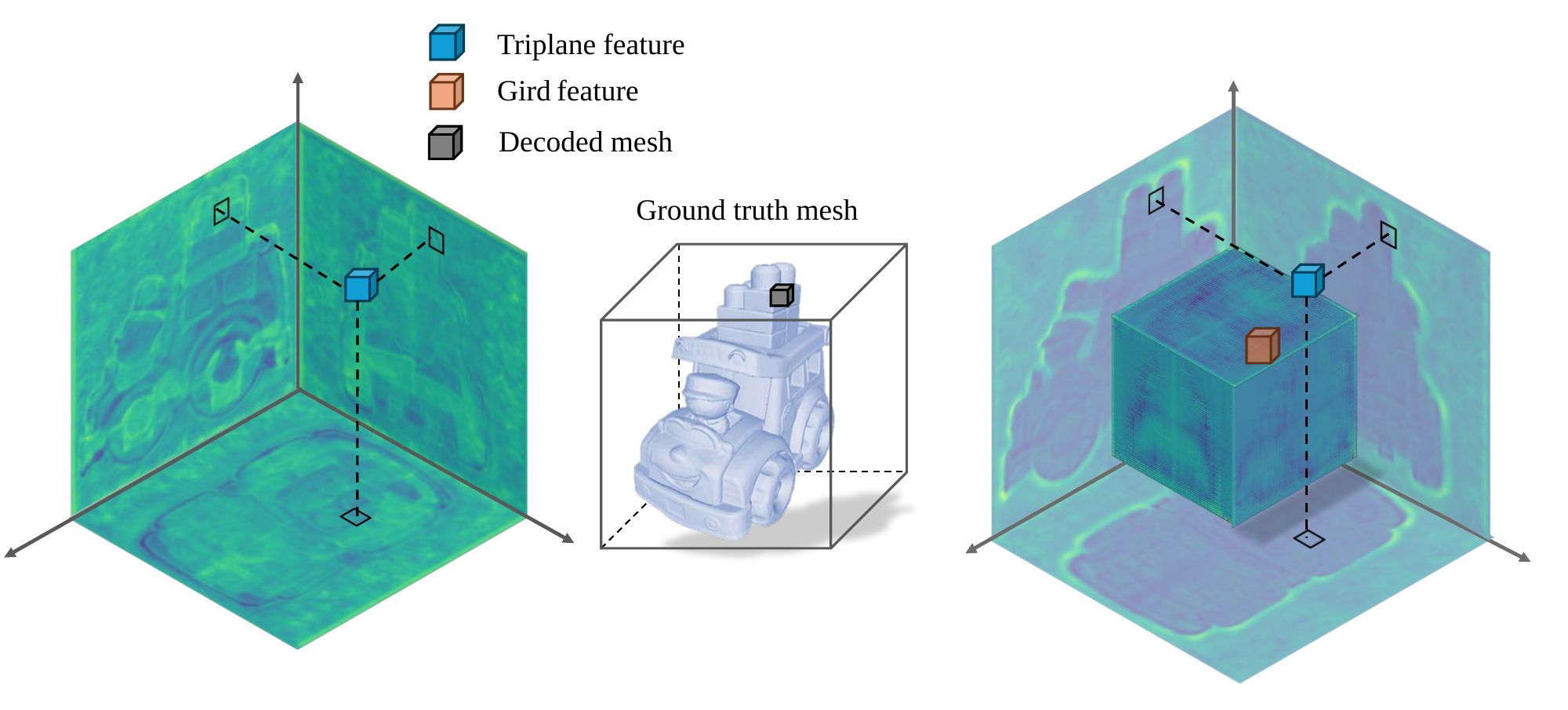}
    \caption{\textbf{Latent space representation for 3D shape VAE: Triplane vs. Hybrid Triplane.} The \textbf{left} depicts the triplane representation composed of three 2D planes. On the \textbf{right}, we propose a hybrid triplane representation that integrates a high-resolution triplane with a low-resolution 3D grid. This design incorporates explicit 3D representation in the latent space while keeping the latent size manageable, ensuring high resolution without excessive complexity or training difficulty.}
    \label{fig:home}
    \vspace{-5mm}
\end{figure}

\begin{figure*}[t]
    \centering
    \includegraphics[width=\textwidth]{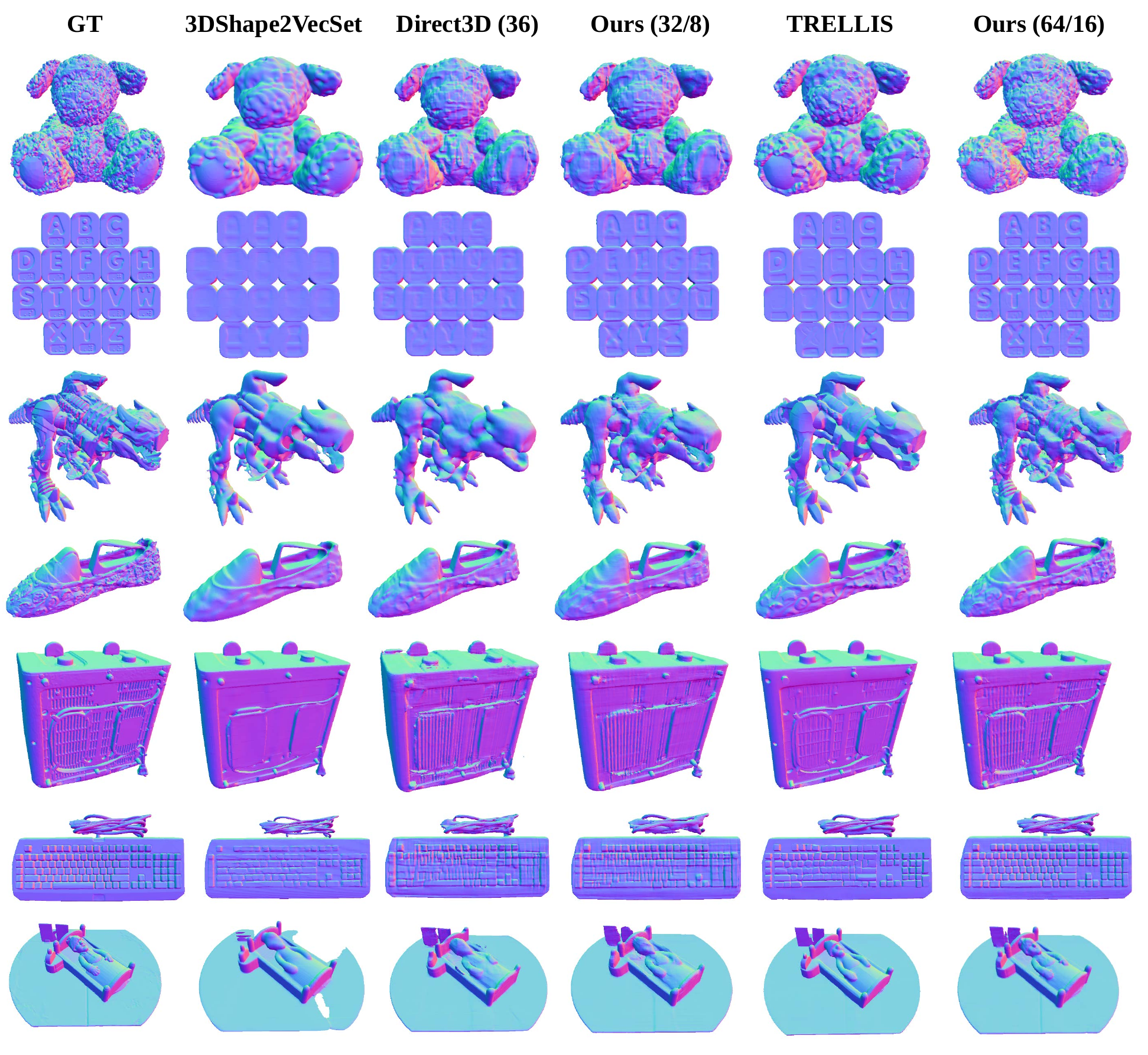}
    \vspace{-6mm}
    \caption{\textbf{Qualitative comparison of VAE reconstruction results.} ``Ours (32/8)" denotes our Hyper3D-VAE, where the resolutions of the latent triplane and latent grid are set to 32 and 8, respectively. ``Direct3D (36)" refers to a variant of Direct3D where we adjust the latent triplane resolution to 36, ensuring a fair comparison between Ours (32/8) and Direct3D (36) with similar latent token lengths (3584 vs. 3888). Similarly, ``Ours (64/16)" represents our model with a latent triplane resolution of 64 and a latent grid resolution of 16, allowing for a comparison with Trellis (16,384 vs. 20,000). (Better viewed with zoom-in.)}
    \label{fig:qualitative_comparison}
    \vspace{-4mm}
\end{figure*}

3D content creation plays a crucial role in enhancing realism and immersion across various domains, such as gaming, filmmaking, and AR/VR applications. However, conventional 3D modeling methods often require extensive technical expertise and manual effort, making them inefficient and less accessible to users without specialized skills. The success of recent AI-powered 3D content generation methods~\cite{hong2024lrmlargereconstructionmodel,camdm,zhang2023temo,liu2023syncdreamer,chen2022tango,lan2024ln3diff,EnVision2023luciddreamer,liu2023one2345,TripoSR2024,cao2024avatargo,cao2024dreamavatar,wang2022rodingenerativemodelsculpting} has reshaped this field by greatly minimizing the manual effort involved in 3D content creation and expanding accessibility for users without prior expertise.

Building on the success and widespread adoption of 2D content generation methods~\cite{ho2020denoising,rombach2022high,zhang2023adding,chen2023pixartalpha,chen2024pixartdelta}, such as text-to-image and image-to-image, 3D content generation approaches aim to extend the capabilities of diffusion models to the 3D domain. 
Early approaches~\cite{chen2023cascade,chen2024microdreamer,gao2024cat3d,liu2023oneplus,liu2023one2345,liu2023zero,liu2023syncdreamer,long2023wonder3d,lu2023direct25,qiu2023richdreamer,zheng2024mvd} in this domain employ a multi-view diffusion model to extend a single-view image into multiple perspectives, followed by 3D shape reconstruction using either sparse-view reconstruction or score distillation sampling (SDS) optimization~\cite{poole2022dreamfusion}. 
While these methods can produce high-quality 3D shapes, they suffer from significant inefficiency due to the computationally intensive nature of sparse reconstruction and SDS optimization. 
Moreover, these pipelines heavily rely on the multi-view diffusion model’s ability to generate multi-view images, yet diffusion models often introduce inconsistencies across different views, further limiting their effectiveness.

To address the limitations of optimization-based methods, recent research~\cite{wu2024direct3d,ren2024xcube,shape2vecset,zhang2024clay} explores an alternative approach that bypasses the indirect generation of multi-view images and instead focuses on directly generating 3D shapes from single-view images using a native 3D diffusion model. These pipelines typically consist of two stages: first, encoding 3D shapes into a latent space using variational autoencoders (VAEs), and second, training a latent diffusion model to generate realistic and high-quality 3D representations. The effectiveness of this generative pipeline largely depends on the VAE’s ability to accurately encode and reconstruct 3D shapes~\cite{li2024craftsman,zhang2024lagem}. In this paper, we aim to enhance the design of 3D VAEs to improve the performance of 3D shape generation.

Unlike 2D image VAEs, where the input image is fully observable, 3D structures cannot be directly used as inputs. Existing 3D VAEs typically address this issue by uniformly sampling points on mesh surfaces for shape encoding and subsequently reconstructing the original 3D meshes through the decoder.
However, such a sampling process fails to capture the intricate geometric details of the surface, thereby compromising the performance of the 3D VAE.
To mitigate this limitation, a recent work, Dora~\cite{chen2024dora}, introduces a non-uniform point sampling strategy based on the surface curvature of 3D shapes. However, this approach relies on manually predefined thresholds to identify salient edges and sharp regions, which may introduce heuristic biases.
Moreover, for 3D VAEs, the choice of latent space representation is crucial, as it determines how effectively a limited number of tokens can encode complex 3D geometric information.
Existing methods, such as 3DShape2VecSet~\cite{shape2vecset}, use a 1D VecSet representation as the latent representation, while Direct3D~\cite{wu2024direct3d} adopts a 2D triplane representation. Notably, these structures cannot serve as explicit 3D representations in the latent space, posing the risk of failing to capture intricate geometric details.
Additionally, recent works, such as Trellis~\cite{xiang2024structured} and MeshFormer~\cite{liu2024meshformer} have begun exploring 3D latent representations by utilizing structured 3D latents or 3D feature volumes. However, due to the large latent size, these approaches struggle to achieve high resolution and face significant computational challenges, requiring complex training strategies.

To address these challenges, we propose Hyper3D, a novel approach that improves VAE reconstruction by introducing an efficient 3D representation that integrates hybrid triplane and octree features.
First, when constructing VAE inputs from a 3D training dataset, we move away from the conventional approach of uniform sampling points on the mesh surface. 
Instead, we leverage a pre-trained octree feature model to embed mesh information into the network. This strategy mitigates the loss of geometric surface details caused by uniform point sampling. Moreover, compared to Dora's~\cite{chen2024dora} sharp edge-based sampling strategy, our approach adaptively extracts fine-grained geometric features from 3D shapes without requiring manually predefined thresholds.
Furthermore, to overcome the shortcomings of existing latent representations, we incorporate low-resolution 3D grid features and integrate them with high-resolution triplane latents to construct a hybrid triplane latent representation (See in \cref{fig:home}). 
The incorporation of 3D grid information compensates for the lack of explicit 3D representations in the VAE latent space, enabling more effective compression and representation of intricate geometric details in 3D shapes.
Moreover, the compact size of the low-resolution grid latent not only reduces computational costs but also simplifies training compared to existing structured 3D latent representations. Meanwhile, its integration with triplane latents ensures the preservation of high-resolution details.

Through extensive experiments, we demonstrate the proposed Hyper3D approach's capability to reconstruct high-quality geometric details and its strong generalization ability. \cref{fig:qualitative_comparison} presents a comparison of 3D-VAE reconstruction performance against existing methods. To summarize, our major contributions include:
\begin{itemize}
    \item We introduce Hyper3D, a novel 3D VAE approach that leverages octree features to refine the conventional uniform sampling method for VAE inputs, significantly preserving the geometric details of shape surfaces.
    \item We introduce a hybrid triplane as the latent space representation for Hyper3D, integrating low-resolution 3D grid and high-resolution triplane features to overcome the limitations of existing methods, which either lack explicit 3D representations in the latent space or struggle to achieve high resolution and efficient training.
    \item Through extensive experiments, we demonstrate that Hyper3D significantly enhances the reconstruction quality of geometric structures as a VAE.
\end{itemize}
\section{Related Work}
\subsection{3D Content Creation}
Existing 3D generation methods can be categorized into three groups. 
\ding{182} Optimization methods~\cite{poole2022dreamfusion,lin2023magic3d, fantasia3d, wang2023prolificdreamer,RichDreamer,sweetdreamer,DreamGaussian, shi2023MVDream,liu2023syncdreamer, liu2023sherpa3d} based on 2D diffusion model priors, pioneered by DreamFusion~\cite{poole2022dreamfusion}, leverage score distillation sampling (SDS) to enhance 3D representations~\cite{mildenhall2020nerf,kerbl20233d, shen2021deep}. While these methods are capable of producing photorealistic results, their reliance on continuously optimizing Neural Radiance Fields (NeRF)~\cite{mildenhall2021nerf} for each object or scene results in slow generation, training instability, and challenges in preserving geometric consistency. 
\ding{183} Large-scale reconstruction models, such as LRM~\cite{LRM} and subsequent studies~\cite{xu2024instantmesh,li2023instant3d, wang2024crm, tang2024lgm,liu2024meshformer,wang2023pf}, utilize sparse-view reconstruction to enhance the efficiency of 3D generation. However, the absence of explicit geometric priors often results in degraded geometric accuracy and inconsistencies in surface details. 
\ding{184} 3D-native generative models~\cite{shape2vecset, li2024craftsman, wu2024direct3d, zhang2024clay,zhao2023michelangelo,xiang2024structured,liu2024meshformer}, exemplified by 3DShape2VecSet~\cite{shape2vecset}, Direct3D~\cite{shuang2025direct3d} and Trellis~\cite{xiang2024structured}, follow a two-stage paradigm: initially, a 3D VAE is trained to encode shapes into a latent space, followed by a conditional latent diffusion model for generation. This methodology leverages the geometric constraints inherent to VAEs, leading to improved geometric consistency. However, the fidelity of generated shapes remains inherently constrained by the VAE’s reconstruction capability. Recent studies~\cite{li2024craftsman,zhang2024lagem} have demonstrated that enhancing VAE reconstruction directly benefits downstream generation quality, underscoring the motivation to refine VAE designs.

\begin{figure*}
    \centering
    \includegraphics[width=\textwidth]{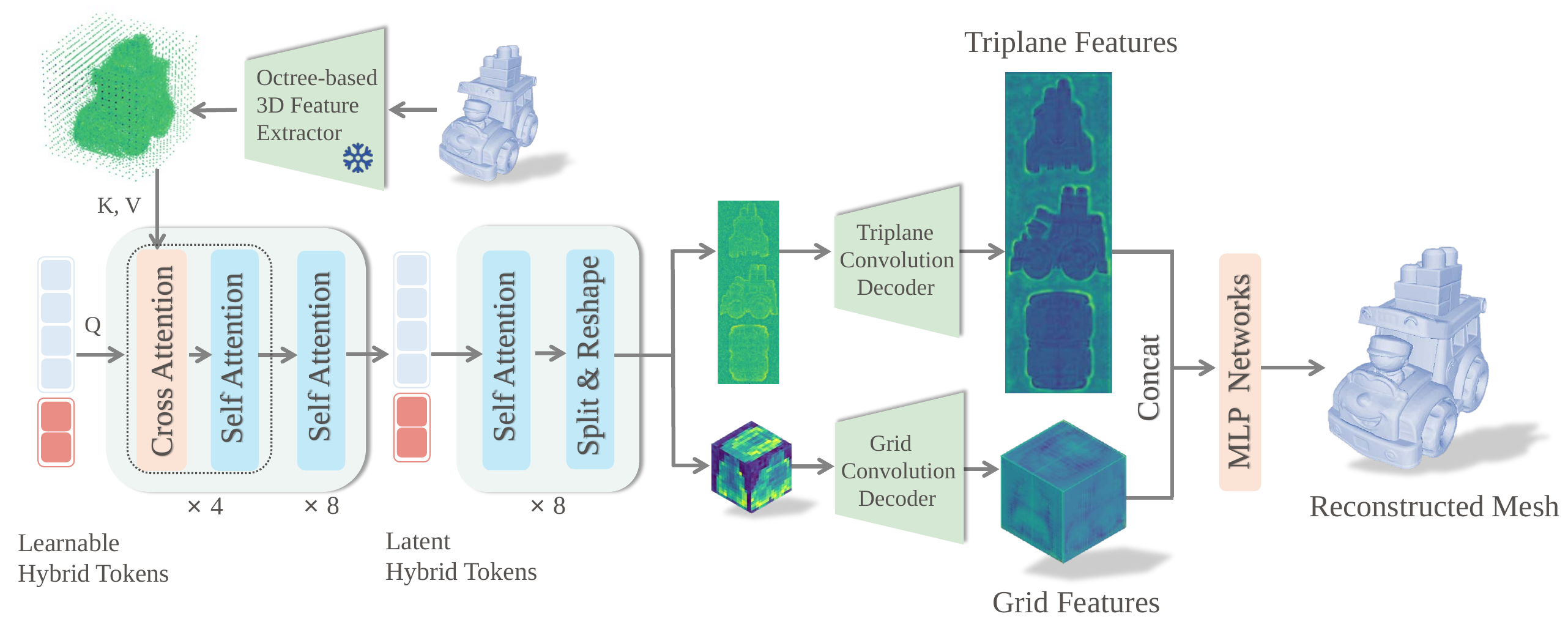}
    \vspace{-8mm}
    \caption{\textbf{Overview of the proposed Hyper3D-VAE.} Instead of relying solely on uniform sampling on mesh surfaces, we utilize an octree-based 3D feature extractor to capture high-frequency geometric details more effectively. During encoding, we introduce learnable triplane tokens and learnable grid tokens, which are concatenated to form learnable hybrid tokens. This design enables the model to effectively capture spatial dependencies across both 2D and 3D representations. In our decoder, the latent hybrid tokens are separated and reshaped into their respective 2D and 3D structures, followed by several upconvolutional layers.}
    \vspace{-4mm}
    \label{fig:method}
\end{figure*}

\subsection{3D Shape Variational Auto-Encoders}
Building a 3D shape VAE necessitates two critical aspects: effectively encoding complex 3D structures into the VAE and selecting an appropriate latent space representation for compressed 3D features. 
Various approaches exist for constructing VAE inputs, one of which is volume-based methods~\cite{xiong2024octfusion, ren2024xcube}. For example, XCube~\cite{ren2024xcube} employs sparse convolution to encode voxelized surfaces, enabling high-fidelity reconstruction. However, these methods rely on large latent representations, often exceeding 10,000 tokens. While they effectively retain geometric details, the high-dimensional latent space presents substantial challenges for training downstream diffusion models.
In contrast, recent studies have explored using uniformly sampled surface points as VAE inputs~\cite{shape2vecset, wu2024direct3d, li2024craftsman, zhang2024clay,zhao2023michelangelo}, as seen in 3DShape2VecSet~\cite{shape2vecset} and Direct3D~\cite{wu2024direct3d}. While this approach significantly reduces the size of the latent representation and accelerates the training of downstream diffusion models, uniform sampling leads to substantial loss of geometric details on the 3D shape’s surface.
Dora~\cite{chen2024dora} introduces a non-uniform point sampling strategy that leverages surface curvature to enhance 3D shape representation. However, this method depends on manually predefined thresholds to identify salient edges and sharp regions, which may introduce heuristic biases and limit its adaptability across diverse geometries.

Beyond input representation, the choice of latent space representation plays a crucial role. 3DShape2VecSet~\cite{shape2vecset} directly compresses sampled points into a 1D implicit latent representation, while Direct3D~\cite{wu2024direct3d} adopts an explicit triplane latent representation. Notably, the triplane representation remains a 2D explicit encoding that approximates 3D structures. The absence of explicitly preserved 3D structural information within the VAE’s latent space leads to significant geometric detail loss in the reconstructed 3D shapes.
Furthermore, explicit 3D structural latents, as employed by Trellis~\cite{xiang2024structured} and MeshFormer~\cite{liu2024meshformer}, enhance shape representation but suffer from high-dimensional latent spaces. This not only complicates the training of diffusion models but also presents challenges in maintaining high-resolution reconstructions.
Thus, achieving a balance between geometric fidelity, latent compactness, and downstream model efficiency remains a fundamental challenge in designing 3D shape VAEs.
\section{Method}

In this section, we introduce how Hyper3D compresses 3D shapes into a compact and efficient latent space while preserving fine-grained 3D structural details. We begin with a brief overview of existing 3D shape VAE approaches in \cref{sec:preliminary}. Subsequently, \cref{sec:octree} and \cref{sec:representation} present octree-based features and Hybrid Triplane as two representations of 3D shapes within our VAE framework. Finally, \cref{sec:VAE} and \cref{sec:scalup} introduce our proposed hybrid triplane-based VAE architecture and the scaling-up scheme.
\subsection{Preliminary: Tiplane-based 3D shape VAE}
\label{sec:preliminary}
Direct3D~\cite{wu2024direct3d} proposes a novel 3D shape VAE based on the triplane representation. For the encoder, the mesh is first discretized into a set of points uniformly sampled on the surface along with their corresponding normal vectors. These points are then passed through a Fourier embedder~\cite{jaegle2021perceiver}, represented as $\mathbf{P} \in \mathbb{R}^{N_P \times C_P}$, where $N_P$ and $C_P$ denote the number of the point cloud sampled from the mesh surface and the channel dimension of the point embeddings, respectively. Then, a set of learnable triplane tokens $\mathbf{e} \in \mathbb{R}^{L_e \times C_e}$ is utilized to query point cloud information through a cross-attention layer. Subsequently, multiple layers of self-attention are applied to facilitate information exchange among tokens at different spatial locations, thereby obtaining the latent representations $\mathbf{z} \in \mathbb{R}^{L_e \times C_z}$.

During the decoding process, the latent representation $\mathbf{z}$ first passes through several layers of self-attention to enhance feature interactions and refine the representation by capturing long-range dependencies. The features are then progressively upsampled through a series of upsampling layers, yielding the final triplane representation \( \mathbf{T} \in \mathbb{R}^{C \times R \times R} \), where $C$ and $R$ denotes the channel dimension and the resolution of decoded triplane.

Finally, the occupancy field \( \hat{O}\) is predicted by an MLP decoder, which takes as input the interpolated features $(\mathbf{f_{XY}}, \mathbf{f_{YZ}}, \mathbf{f_{XZ}})$ from \( \mathbf{T} \).

The VAE loss function is defined as:
\begin{equation}
\mathcal{L}_{\text{VAE}} = \mathcal{L}_{\text{BCE}} + \lambda \mathcal{L}_{\text{KL}},
\end{equation}
where $\mathcal{L}_{\text{BCE}}$ is the Binary Cross-Entropy (BCE) reconstruction loss and $\mathcal{L}_{\text{KL}}$ is the KL divergence regularization term.
Although this approach enables the generation of 2D latent representations for diffusion, its use of a uniform sampling strategy constrains its ability to capture fine geometric details. Additionally, the 2D triplane representation lacks an explicit 3D structure to effectively encode spatial information. To overcome these limitations, our work introduces a novel input strategy and a hybrid 3D representation.

\subsection{Octree-based Feature Extractor}
\label{sec:octree}
Existing 3D shape VAEs~\cite{zhang2024clay, shape2vecset, wu2024direct3d} predominantly employ uniform sampling strategies to incorporate 3D shape surface information into the network. However, such uniform sampling often fails to capture complex geometric structures, particularly in sharp-edge regions. We propose a novel approach to integrate raw 3D shape information into the VAE, which achieves higher geometric precision with fewer input points. Octree-based representations have been used in 3D shape processing due to their hierarchical structure~\cite{ohtake2005multi, wang2022dual, xiong2024octfusion}, which efficiently captures both global and local geometric details. Compared to the uniform sampling strategy, octree-based methods adaptively allocate higher resolution to regions with rich geometric features, leading to improved reconstruction fidelity. However, prior works~\cite{wang2017cnn, wang2022dual, xiong2024octfusion} have shown that diffusion models based on octree representations often require complex training strategies, such as cascaded training and the simultaneous generation of both the splitting status and the latent features of each octree node, where high precision is required at each stage of generation. Motivated by these insights, we leverage octree features as the input of our VES rather than a latent representation for diffusion, as they provide rich geometric details while avoiding the challenges associated with direct generative modeling. 

We adopt an octree-based feature extractor proposed by~\cite{xiong2024octfusion} to extract octree features from 3D shapes. For each 3D shape, its octree structure \( \mathcal{O} \) is first constructed based on the surface distribution of the shape. Given an octree resolution of \( 2^n \), each leaf node \( \mathcal{L}_i \) stores both the splitting information and the signed distance function (SDF) value. An octree-based feature extractor \( \mathcal{F} \) is trained under the supervision of the information stored at each leaf node across different levels. Given a 3D shape \( \mathbf{S}\), we extract the features from the leaf nodes at the $l_{th}$ level as the input to our VAE model:
\begin{equation}
\mathbf{P}_{oct} = \mathcal{F}^{(l)}(\mathcal{O}_S),
\end{equation}
where $\mathbf{P}_{oct} \in \mathbb{R}^{N_{oct} \times C_{oct}}$, $\mathcal{O}_{S}$ denotes the octree structure of the 3D shape. $N_{oct}$ and $C_{oct}$ denote the number of leaf nodes and octree feature channels, respectively.

The octree features capture hierarchical spatial geometry details of the 3D shape at different space resolutions. For further details on the architecture and training of the octree feature extractor, please refer to our appendix.

\subsection{Hybrid Triplane Representation}
\label{sec:representation}
Triplane representation compresses 3D shape information into three 2D feature planes. However, this 2D-to-3D approximation inevitably leads to the loss of 3D structured information. Traditional 3D representations~\cite{xiang2024structured,liu2024meshformer}, despite their explicit characteristics, suffer from higher computational costs and require intricate training strategies.

Inspired by MERF~\cite{reiser2023merf}, we propose a hybrid triplane representation for 3D shape reconstruction. As illustrated by \cref{fig:home}, our proposed hybrid triplane consists of a triplane \( \mathbf{T} \in \mathbb{R}^{C \times R \times R} \) and a grid $\mathbf{G} \in \mathbb{R}^{C \times R_G \times R_G \times R_G}$,  represented as [${\mathbf{T}, \mathbf{G}}$]. Given an arbitrary query point $\mathbf{q} = (x, y, z)$ in space, we first obtain its triplane feature $(\mathbf{f_{XY}}, \mathbf{f_{YZ}}, \mathbf{f_{XZ}})$ from a high-resolution triplane via bilinear interpolation from the $\mathbf{T_{XY}}$
, $\mathbf{T_{YZ}}$, and $\mathbf{T_{XZ}}$ planes.
Simultaneously, a 3D-structured feature $\mathbf{g}$ is queried from a low-resolution 3D grid $\mathbf{G}$. These features are concatenated to form the final hybrid triplane feature $\mathbf{F}_q$ of point $\mathbf{q}$:
\begin{equation}
\mathbf{F}_q = \text{Concat}(\mathbf{f_{XY}}, \mathbf{f_{YZ}}, \mathbf{f_{XZ}}, \mathbf{g}),
\label{eq:feature_concat}
\end{equation}
This hybrid representation captures fine-grained 2D details at high resolution $R$ while maintaining global 3D structural information at lower resolution $R_G$.

\subsection{Architecture of Hyper3D-VAE}
\label{sec:VAE}
\cref{fig:method} illustrates the overall framework of our proposed VAE. Given an input 3D shape \( \mathbf{S} \), we first use the pre-trained octree feature extractor \( \mathcal{F} \) to get octree features. 
The coordinates of the leaf nodes are then Fourier-embedded and concatenated with octree features. 
Our VAE employs learnable tokens \( \mathbf{e} \), including grid tokens \( \mathbf{e}_{G} \) and triplane tokens \( \mathbf{e}_{T} \), learning spatial information at lower and higher resolutions, respectively. 
A series of attention blocks, each containing a cross-attention layer and a self-attention layer, are utilized to enhance 3D shape feature injection. 
Subsequent self-attention layers facilitate interaction between triplane and grid tokens, encoding the shape into latent triplane tokens \( \mathbf{z}_{T} \) and latent grid tokens \( \mathbf{z}_{G} \).
The encoding process can be formulated as:
\begin{equation}
\mathbf{P} = \operatorname{Concat}(\operatorname{Fourier}(\mathbf{S}), \mathcal{F}^{(l)}(\mathcal{O}_S)),
\end{equation}
\begin{equation}
\mathbf{e}' = \operatorname{SelfAttn}(\operatorname{CrossAttn}(\mathbf{e}, \mathbf{P})),   \mathbf{z} = \operatorname{SelfAttn}(\mathbf{e}'),
\end{equation}
where $\mathbf{e}'$ denotes the latent features. For decoding, these latent tokens $\mathbf{z}$ undergo self-attention layers and are then reshaped into respective resolutions 
\( \mathbf{z}_{T} \in \mathbb{R}^{C_{z} \times 3r_{T} \times r_{T}} \),
\( \mathbf{z}_{G} \in \mathbb{R}^{C_{z} \times r_{G} \times r_{G} \times r_{G}} \).
Subsequently, to decode the latent features into high-resolution features, they are processed by a series of 2D and 3D upconvolutional layers \( \operatorname{UpConv2D} \) and \( \operatorname{UpConv3D} \), respectively. We can obtain the final triplane $\mathbf{T}$ and grid features $\mathbf{G}$ with resolutions of \( R_{T} \) and \( R_{G} \).
The decoder can be formulated as:
\begin{equation}
\mathbf{z}_{T}, \mathbf{z}_{G} = \operatorname{Reshape}(\operatorname{SelfAttn}(\mathbf{z})),
\end{equation}
\begin{equation}
\mathbf{T} = \operatorname{UpConv2D}(\mathbf{z}_{T}), \quad \mathbf{G} = \operatorname{UpConv3D}(\mathbf{z}_{G}).
\end{equation}
A geometry MLP then decodes the concatenated feature obtained from \cref{eq:feature_concat} for each query point into occupancy values. Once the occupancy field is obtained, it can be transformed into a triangle mesh with the Marching cubes algorithm~\cite{lorensen1998marching}. Inspired by Direct3D~\cite{wu2024direct3d}, we employ semi-continuous occupancy as supervision for more stable training and finer surface details.

\subsection{Resolution Upscaling Strategy}
\label{sec:scalup}
The initialization of learnable tokens plays a crucial role in efficiently upscaling the resolution of our Hyper3D-VAE. Given a trained low-resolution model with the learnable tokens \( \mathbf{e} \in \mathbb{R}^{L_e \times C_e} \), it is first reshaped into the size of  $\mathbf{z}_{T}, \mathbf{z}_{G}$ and then passed through a parameter-free 2D and 3D upsampling layer to achieve resolution upscaling initialization as well as faster convergence at higher resolutions.

\section{Experiment}
\subsection{Implement details}

We utilize a filtered subset of the Objaverse dataset~\cite{deitke2023objaverse} containing 160,000 3D objects as our training data. The selection is based on the average aesthetic score computed from four rendered views of each object. For details regarding the aesthetic assessment system, please refer to \cite{xiang2024structured}. Each 3D model is normalized to fit within a unit sphere centered at the world origin. We then follow CLAY~\cite{zhang2024clay} to preprocess all meshes into watertight geometries.

Our VAE takes as input $N_{oct} = 30,720$ leaf node points along with their corresponding octree features. The octree features are extracted from an octree structure with depth $l = 6$ and a channel dimension of 64. The encoder consists of 4 alternating cross-attention and self-attention layers, followed by 8 self-attention layers. For the decoder, we employ 8 self-attention layers followed by 3 upconvolutional layers to progressively upsample both the triplane and grid representations. We apply four random perturbations to each octree leaf node to construct our supervision point set. During training, we use 40,960 supervision points to compute the loss.

The VAE model is trained on 32 H20 GPUs for 4 days, with the batch size varying from 16 to 80 per GPU, depending on the model size. The KL regularization weight is set to $\lambda = 1 \times 10^{-4}$. We employ the AdamW optimizer \cite{loshchilov2017adamw} with a learning rate of $1 \times 10^{-4}$.

\definecolor{best}{rgb}{1.0, 0.6, 0.6}   
\definecolor{second}{rgb}{1.0, 0.8, 0.6} 
\definecolor{third}{rgb}{1.0, 1.0, 0.6}  

\begin{table}[t]
    \centering
    \renewcommand{\arraystretch}{1.2} 
    \resizebox{\linewidth}{!}{
    \begin{tabular}{l|>{\centering\arraybackslash}p{2.1cm} >{\centering\arraybackslash}p{2.1cm} >{\centering\arraybackslash}p{2.1cm} >{\centering\arraybackslash}p{2.1cm}}
        \toprule
        Model &  $\uparrow$ F-Score & $\downarrow$ CD  $\times$ 10000 & $\uparrow$ NC & $\uparrow$ Surface IoU \\
        \midrule
        ours (32/4) & \cellcolor{second} 0.8601 & \cellcolor{second}  14.1703 & \cellcolor{third} 0.8954 & \cellcolor{best} 0.5984 \\
        \cellcolor{gray! 40}ours (32/8) & \cellcolor{best} 0.9267 & \cellcolor{best} 7.4873 & \cellcolor{best} 0.9251 & \cellcolor{second} 0.5769 \\
        ours (32/16) & \cellcolor{third} 0.8349 & \cellcolor{third} 16.3317 & \cellcolor{second}  0.8957 & \cellcolor{third} 0.4651 \\
        \bottomrule
    \end{tabular}
    }
\vspace{-2mm}
\caption{\textbf{Quantitative comparison of different grid resolutions in hybrid triplane representation.} The \textcolor{best}{red}, \textcolor{second}{orange}, and \textcolor{yellow}{yellow} colors denote the best, second-best, and third-best results, respectively. The \textcolor{gray! 40}{gray} color means settings of our Hyper3D-VAE model. The following tables follow the same pattern.}
\vspace{-5mm}
\label{tab:ablation_grid}
\end{table}

\definecolor{best}{rgb}{1.0, 0.6, 0.6}   
\definecolor{second}{rgb}{1.0, 0.8, 0.6} 
\definecolor{third}{rgb}{1.0, 1.0, 0.6}  

\begin{table*}[t]
    \centering
    \renewcommand{\arraystretch}{1.2} 
    \resizebox{\linewidth}{!}{
    \begin{tabular}{l| c| c | >{\centering\arraybackslash}p{2.1cm} >{\centering\arraybackslash}p{2.1cm} >{\centering\arraybackslash}p{2.1cm} >{\centering\arraybackslash}p{2.1cm}}  
        \toprule
        Method & Latent Representation & Latent Token Length & $\uparrow$ F-Score & $\downarrow$ CD  $\times$ 10000 & $\uparrow$ NC & $\uparrow$ Surface IoU \\
        \midrule
        3DShape2VecSet & Vector set & 4096 / 8192 & 0.9950 & 7.2665	& 0.9343 & 0.4547 \\
        Direct3D (36)	& Triplane	& 3888	& 0.9783 & 21.1667 &	0.9473	& \cellcolor{third} 0.6375 \\
        Trellis & Sparse Voxel	&  $\approx$ 20000 & \cellcolor{third}0.9958 & \cellcolor{third} 6.6528	& \cellcolor{second} 0.9584	& 0.6191\\
        
        \cellcolor{gray! 40}Ours (32/8) & \cellcolor{gray! 40} Hybrid Triplane & \cellcolor{gray! 40} 3584 & \cellcolor{best} 0.9987 & \cellcolor{second} 5.2716 & 	\cellcolor{third} 0.9572	& \cellcolor{second} 0.6812 \\
        \cellcolor{gray! 40}Ours (64/16) & \cellcolor{gray! 40} Hybrid Triplane	& \cellcolor{gray! 40} 16384	& \cellcolor{best} 0.9987 & \cellcolor{best} 5.0842 & \cellcolor{best} 0.9650	& \cellcolor{best} 0.8331 \\
        \bottomrule
    \end{tabular}
    }
\vspace{-2mm}
\caption{\textbf{Quantitative comparison of different 3D shape VAE.} In addition to the evaluation metrics, we also present the Latent Representations adopted by each VAE method and their Latent Token Lengths.}
\vspace{-5mm}
\label{tab:comparison}
\end{table*}

\subsection{Evaluation}

\textbf{Baselines.} To assess the reconstruction quality of our VAE, we compare it against state-of-the-art open-source models, including 3DShape2VecSet~\cite{shape2vecset}, Direct3D~\cite{wu2024direct3d}, and Trellis~\cite{xiang2024structured}. 3DShape2VecSet and Direct3D do not provide publicly available datasets or model weights. To ensure a fair comparison, we train both baseline models and ours using the same 160,000 Objaverse dataset. 3DShape2VecSet is trained with a Farthest Point Sampling (FPS)~\cite{moenning2003fast} strategy using two latent token lengths of 4096 and 8192. Direct3D is trained with a latent triplane at resolutions of 36 and 74, allowing a fair comparison with our hybrid triplane representation by maintaining a similar latent space size. Trellis provides a model checkpoint trained on 500,000 objects, enabling us to make an inference-based evaluation.

\noindent \textbf{Metrics.} Before the evaluation, all decoded meshes and ground-truth meshes are normalized to the range $[-1,1]$. The following metrics are used for evaluation: 1) F-score, which measures reconstruction accuracy by computing the precision and recall of point correspondences within a small distance threshold. In our experiments, we set the threshold to 0.05. 2) Chamfer Distance (CD), which computes the average distance between each reconstructed point and its nearest ground truth point, capturing the overall shape similarity. 3) Normal Consistency (NC), which evaluates surface smoothness by measuring the cosine similarity between normal vectors of corresponding points on the reconstructed and ground-truth meshes. 4) Surface Intersection over Union (Surface IoU), which assesses surface reconstruction quality by computing the overlap between the predicted and ground-truth near-surface regions. To evaluate our Hyper-VAE model, we randomly sample 60 objects from the Google Scanned Objects (GSO) dataset~\cite{downs2022google}.

\begin{figure}[t]
    \centering
    \includegraphics[width=0.9\linewidth]{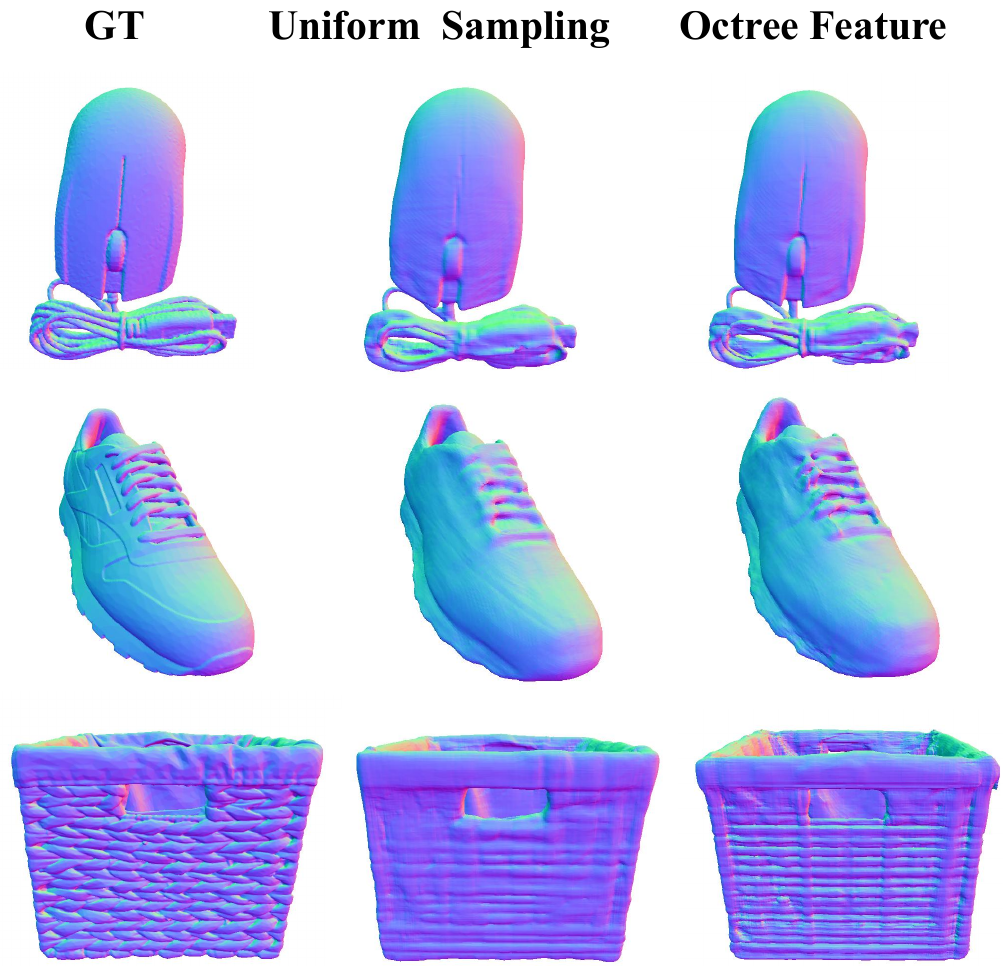}
    \vspace{-2mm}
    \caption{Qualitative comparison for the ablation of VAE with different input strategies.}
    \label{fig:ablation_octree}
    \vspace{-5mm}
\end{figure}

\subsection{Settings for hybrid triplane}
In this section, we conduct experiments on choosing the grid resolution of our hybrid triplane representation. We fix the latent triplane resolution at 32 and vary the grid resolution among 4, 8, and 16. As illustrated in \cref{tab:ablation_grid}, compared to a grid resolution of 8, a lower grid resolution of 4 fails to capture meaningful 3D structured information, whereas a higher grid resolution of 16 results in a large latent token length of 4096 (compared to 3072 for triplane), leading to increased computational cost and degraded performance. Thus, we select a grid resolution of 8 as the optimal setting and use a resolution of 16 when scaling up our model to a higher latent triplane resolution of 64.

\begin{figure}[t]
    \centering
    \vspace{-2mm}
    \includegraphics[width=0.97\linewidth]{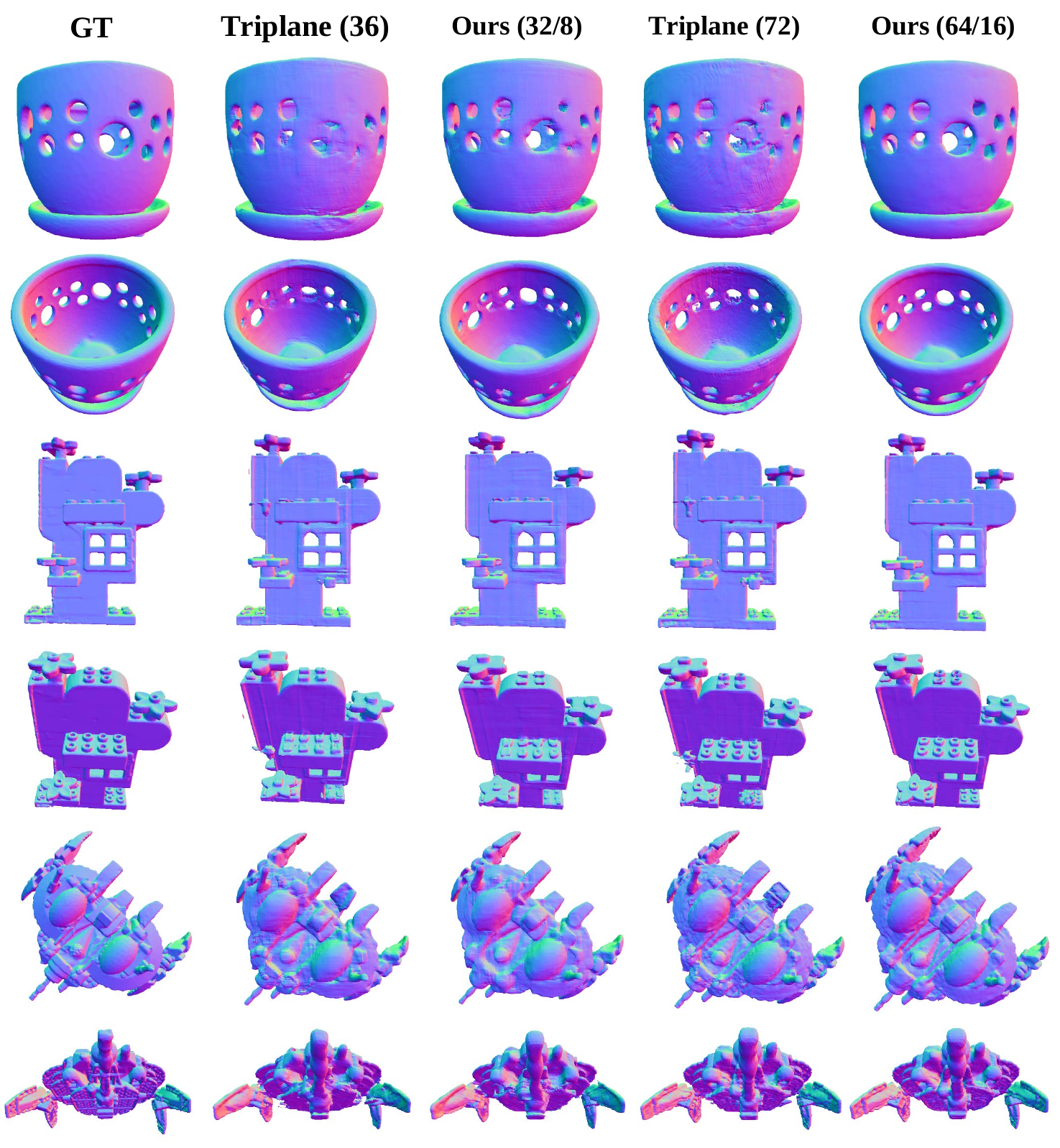}
    \vspace{-2mm}
    \caption{\textbf{Qualitative comparison for the ablation of VAE with different representations.} The second and third columns present a comparison between triplane and hybrid triplane under latent token lengths of 3,888 and 3,584, respectively. The fourth and fifth columns present a comparison under latent token lengths of 16,428 and 16,384, respectively.}
    \label{fig:resolution_ablation_vis}
    \vspace{-5mm}
\end{figure}

\begin{figure}[t]
    \centering
    \includegraphics[width=0.93\linewidth]{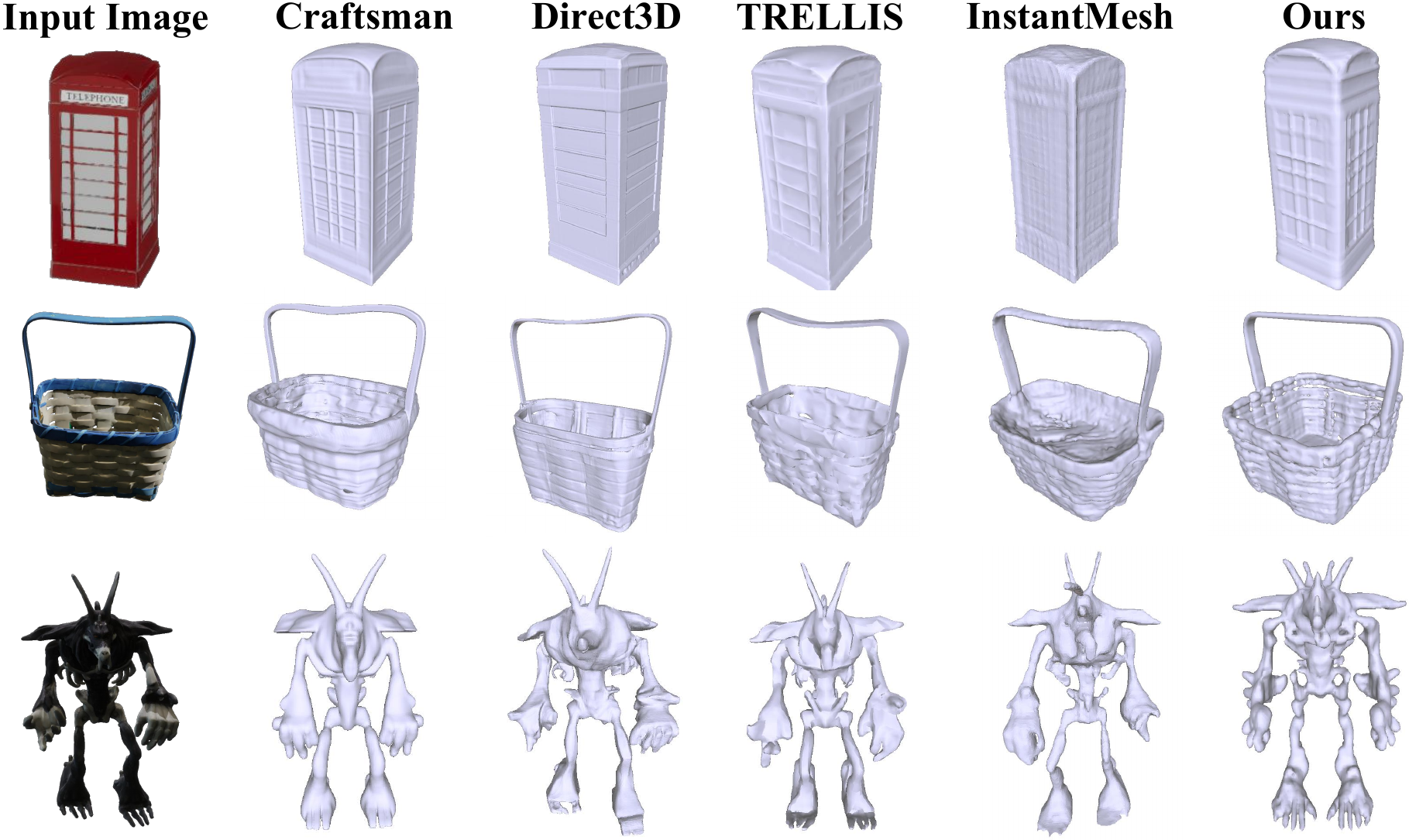}
    \vspace{-2mm}
    \caption{\textbf{DiT results of the image-to-3D generation trained on our Hyper3D-VAE (32/8).} Even when trained on a relatively small dataset of 46,000 objects, the DiT model built upon our proposed Hyper3D generates 3D shapes with fine details and achieves competitive generative performance among the state-of-the-art generation methods.}
    \label{fig:generation_vis}
    \vspace{-5mm}
\end{figure}

\subsection{Quantitative and Qualitative Comparison}

\definecolor{best}{rgb}{1.0, 0.6, 0.6}   
\definecolor{second}{rgb}{1.0, 0.8, 0.6} 
\definecolor{third}{rgb}{1.0, 1.0, 0.6}  

\begin{table*}[t]
    \centering
    \renewcommand{\arraystretch}{1.2} 
    \resizebox{\linewidth}{!}{
    \begin{tabular}{l|c|>{\centering\arraybackslash}p{2.8cm} >{\centering\arraybackslash}p{2.8cm} >{\centering\arraybackslash}p{2.8cm} >{\centering\arraybackslash}p{2.8cm}}
        \toprule
        VAE Input Strategy & \# of input points &  $\uparrow$ F-Score & $\downarrow$ CD  $\times$ 10000 & $\uparrow$ NC & $\uparrow$ Surface IoU \\
        \midrule
        Uniform Sampling & 81920 & \cellcolor{second} 0.9931 & \cellcolor{second} 9.5056	& \cellcolor{second} 0.9529 & \cellcolor{second} 0.5632\\
        \cellcolor{gray! 40}Octree feature & \cellcolor{gray! 40}30720 & \cellcolor{best} 0.9969	& \cellcolor{best} 5.7283 & \cellcolor{best} 0.9537 & \cellcolor{best} 0.6502 \\
        \bottomrule
    \end{tabular}
    }
\vspace{-2mm}
\caption{\textbf{Quantitative comparison of different VAE input strategies.} The input points number of each strategy and evaluation metrics are reported.}
\vspace{-4mm}
\label{tab:input}
\end{table*}

\definecolor{best}{rgb}{1.0, 0.6, 0.6}   
\definecolor{second}{rgb}{1.0, 0.8, 0.6} 
\definecolor{third}{rgb}{1.0, 1.0, 0.6}  

\begin{table*}[t]
    \centering
    \renewcommand{\arraystretch}{1.2} 
    
    \resizebox{\linewidth}{!}{
    \begin{tabular}{l| c c c | >{\centering\arraybackslash}p{3cm} >{\centering\arraybackslash}p{3cm} >{\centering\arraybackslash}p{3cm} >{\centering\arraybackslash}p{3cm}}  
        \toprule
        Latent Representation & Triplane Resolution & Gird Resolution & Latent Token Length & $\uparrow$ F-Score & $\downarrow$ CD  $\times$ 10000 & $\uparrow$ NC & $\uparrow$ Surface IoU \\
        \midrule
        Triplane & 36 & - & 3888 & 0.9783 & \cellcolor{third} 21.1667 & 0.9473 & 0.6375 \\
        \cellcolor{gray!40}Hybrid Triplane & \cellcolor{gray!40}32 & \cellcolor{gray!40}8 & \cellcolor{gray!40}3584 & \cellcolor{best} 0.9987 & \cellcolor{second} 5.2716 & \cellcolor{third} 0.9572 & \cellcolor{third} 0.6812 \\
        Triplane & 74 & - & 16428 & \cellcolor{third} 0.9790 & 25.2078 & \cellcolor{third} 0.9540 & \cellcolor{second} 0.7624 \\
        \cellcolor{gray!40}Hybrid Triplane & \cellcolor{gray!40}64 & \cellcolor{gray!40}16 & \cellcolor{gray!40}16384 & \cellcolor{best} 0.9987 & \cellcolor{best} 5.0842 & \cellcolor{best} 0.9650 & \cellcolor{best} 0.8331 \\
        \bottomrule
    \end{tabular}
    }
\vspace{-2mm}
\caption{\textbf{Quantitative comparison of VAE representations between naive triplane and hybrid triplane.} Under the same experimental settings, we use different model's latent representation to compare the reconstruction performance under the same latent token length. We present the resolution of the  latent triplane (and grid) representation in each experiment and its total length (i.e., the latent token length.)} 
\vspace{-4mm}
\label{tab:hybrid}
\end{table*}

\cref{tab:comparison} presents the comparison of our Hyper3D-VAE with several baselines across different evaluation metrics. We showcase our VAE with a hybrid triplane representation under two latent sizes to ensure a fair comparison with other methods. As observed from \cref{tab:comparison}, our VAE achieves superior reconstruction quality compared to 3DShape2VecSet and Direct3D with a lower latent token length of 3584 (corresponding to a latent triplane resolution of 32 and grid resolution of 8). 

For a fair comparison with Trellis, which utilizes an active latent grid token length of approximately 20,000, we scale up our model to a latent token length of approximately 16,000 (corresponding to a latent triplane resolution of 64 and a grid resolution of 16). Our scaled-up model only takes 1 day to train on 16 H20 GPUs. Despite using fewer training data (160k vs. 500k) and a smaller latent token length, our Hyper3D-VAE improves the Surface IoU from 0.7624 to 0.8331 compared to Trellis VAE.

\cref{fig:qualitative_comparison} illustrates qualitative comparisons among different VAEs. As observed, 3DShape2VecSet tends to reconstruct smooth surfaces but lacks fine details. Compared to Direct3D (36), our VAE (32/8) reconstructs more details while reducing floaters due to the introduction of the grid representation. Our VAE (64/16) achieves higher-resolution reconstruction than Trellis, leading to superior reconstruction of high-detail surfaces.

\subsection{Ablation Study}

\textbf{Efficacy of Octree Features.} We conduct an ablation study on the input of our VAE using a subset of 46,000 objects from Objaverse. As shown in \cref{tab:input} and \cref{fig:ablation_octree}, octree features capture more surface details compared to traditional uniform sampling strategies used by~\cite{shape2vecset,wu2024direct3d}, leading to improved reconstruction quality.

\noindent \textbf{Efficacy of Hybrid Triplane.} To demonstrate the superiority of our hybrid triplane representation over the naive triplane used by Direct3D~\cite{wu2024direct3d}, we conduct a detailed ablation study. As shown in \cref{tab:hybrid}, we train VAE models using both representations while keeping the latent token length comparable. Even with a lower latent token length, our hybrid triplane representation consistently outperforms the naive triplane across all evaluation metrics. 
\cref{fig:resolution_ablation_vis} further visualizes the improvements brought by the hybrid triplane. The introduction of the grid enables better reconstruction of geometric structures. In contrast, the naive triplane struggles to improve quality with higher resolution, leading to rough surfaces and floaters.
\subsection{Application on Image-to-3D Task}
To demonstrate the applicability of our VAE in 3D generation tasks, we follow the setup of Direct3D~\cite{wu2024direct3d} and train a Diffusion Transformer (DiT)~\cite{peebles2023dit} to map images to latent representations. Due to computational constraints, we conduct experiments on a 46,000-object subset of Objaverse using Hyper3D-VAE (32/8) to extract latents. Each object is rendered into 15 random views, and the model is trained on 32 H20 GPUs for 7 days.  \cref{fig:generation_vis} presents qualitative comparisons with state-of-the-art methods (Craftsman~\cite{li2024craftsman}, Direct3D~\cite{wu2024direct3d}, TRELLIS~\cite{xiang2024structured}, InstantMesh~\cite{xu2024instantmesh}), where our work exhibits competitive generation quality.

\section{Conculusion}
In this paper, we introduce Hyper3D, a novel 3D VAE framework that enhances reconstruction quality by leveraging our proposed efficient 3D representation, which integrates hybrid triplane and octree features. First, we refine the conventional uniform sampling method for VAE inputs by incorporating octree features, effectively mitigating the loss of geometric details in shape surfaces. Furthermore, we introduce a hybrid triplane as the latent representation, which consists of a low-resolution 3D grid and a high-resolution triplane. This design addresses the limitations of existing methods that rely on 1D or 2D latent representations, which fail to preserve explicit 3D structures, while also overcoming the challenges of structured 3D latents that suffer from excessive computational complexity and limited resolution. Extensive experiments demonstrate that Hyper3D significantly improves the reconstruction quality of geometric details in 3D shape generation, validating the effectiveness of our proposed efficient 3D representation.

{
    \small
    \bibliographystyle{ieeenat_fullname}
    \bibliography{main}
}

\clearpage
\appendix
\setcounter{figure}{0}
\setcounter{table}{0}
\renewcommand{\thetable}{A\arabic{table}}
\renewcommand{\thefigure}{A\arabic{figure}}


\section{VAE Architecture and Training}
For the encoder, each attention layer consists of 16 heads with a dimension of 64. The latent representation has a channel number of 16. The decoder employs 16 heads with a dimension of 32, followed by 5 ResNet~\cite{he2016resnet} blocks and 3 upconvolution layers for triplane and grid upsampling. The geometric mapping network consists of 5 linear layers with a hidden dimension of 64. Empirically, we apply semi-continuous occupancy in the range of $[-0.02, 0.02]$ near the surface. A larger range increases training difficulty, while a smaller range leads to staircase-like artifacts.

\section{Octree Feature Extractor Training}
We train the octree feature extractor on a 20k subset of Objaverse to learn features at leaf nodes from levels 6 to 9. The average number of leaf nodes is approximately 15k at level 6 and 500k at level 9. The model is trained on 8 H20 GPUs for 7 days to serve as an embedding module for 3D ground truth shapes. For details of model architecture and hyper-parameters, please refer to~\cite{xiong2024octfusion} and its open-source code.

\section{Visualization of Triplane and Hybrid Triplane}
\begin{figure}[t]
    \centering
    \includegraphics[width=0.45\textwidth]{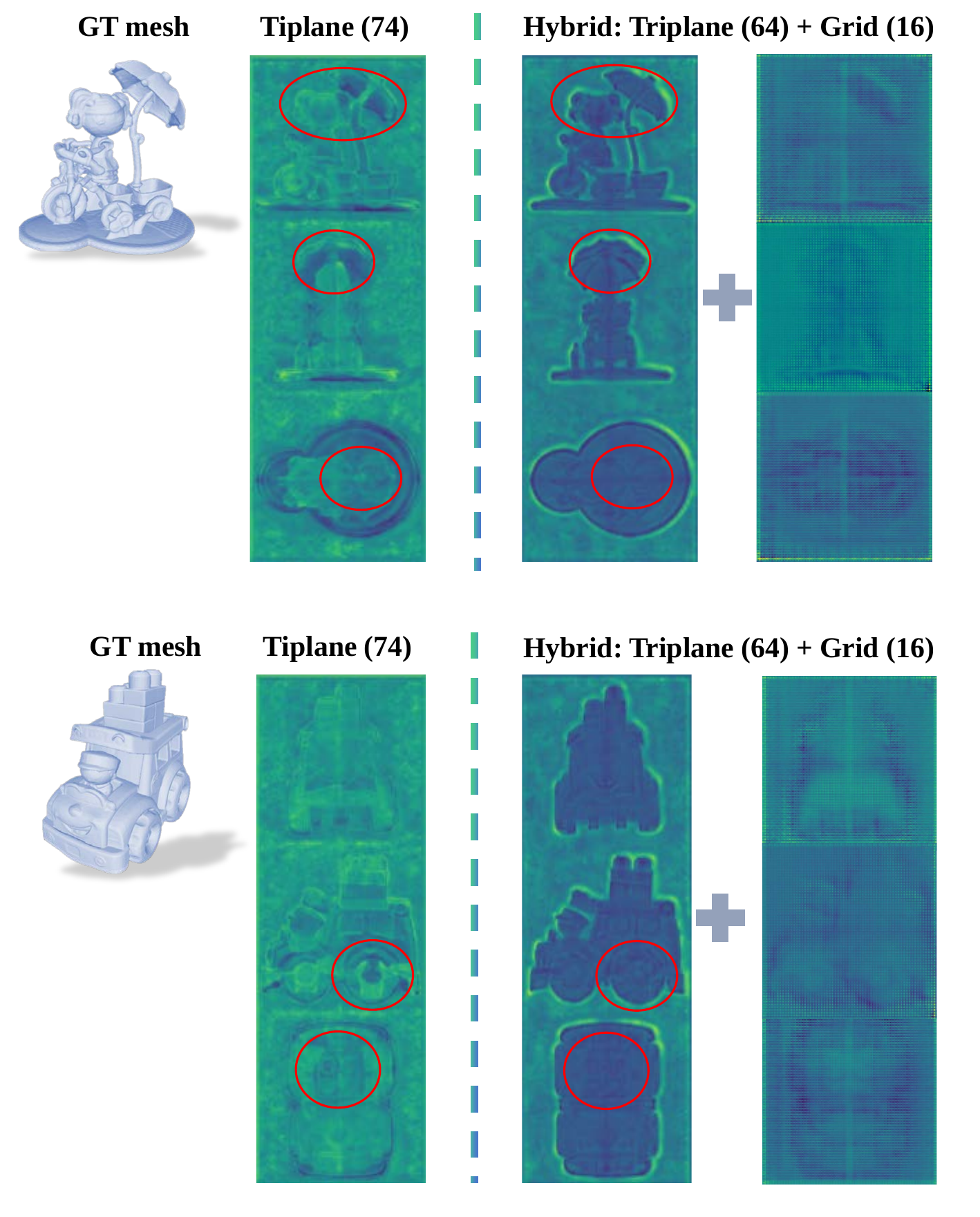}
    \vspace{-2mm}
    \caption{Visualization of decoded triplane vs. decoded hybrid triplane. We make the comparison under the same settings, including VAE input, latent token length, and training data.}
    \vspace{-4mm}
    \label{fig:hybrid}
\end{figure}

To better understand how low-resolution grids contribute to hybrid representation learning, we visualize the decoded triplane and hybrid triplane features in two examples. Each token's color represents the average value along the channel dimension, and the grid is visualized by averaging token features across three coordinate planes. As shown in \cref{fig:hybrid}, the low-resolution grid captures coarse 3D structural information, while the high-resolution triplane captures fine details. The introduction of the grid does not compromise the triplane's representation capacity compared to the naive triplane. Instead, it enhances surface detail learning, as illustrated in the red-circled regions (better viewed by zooming in).

\section{More visualizations}
\begin{figure*}[t]
    \centering
    \includegraphics[width=0.88\textwidth]{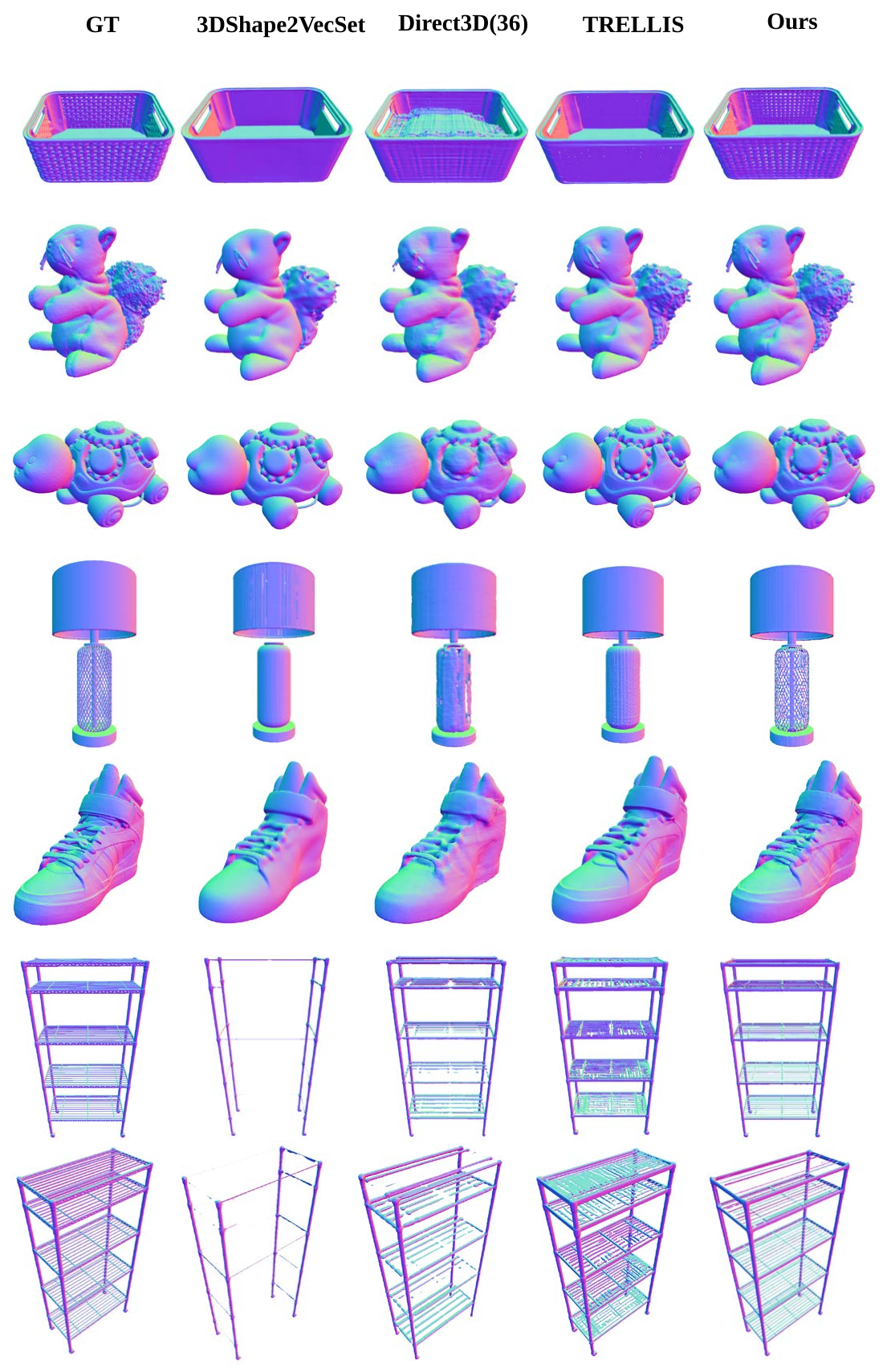}
    \vspace{-2mm}
    \caption{More comparison with the state-of-the-art VAE mdoels.}
    \vspace{-4mm}
    \label{fig:more_vis_appendix}
\end{figure*}

\begin{figure*}[t]
    \centering
    \includegraphics[width=\textwidth]{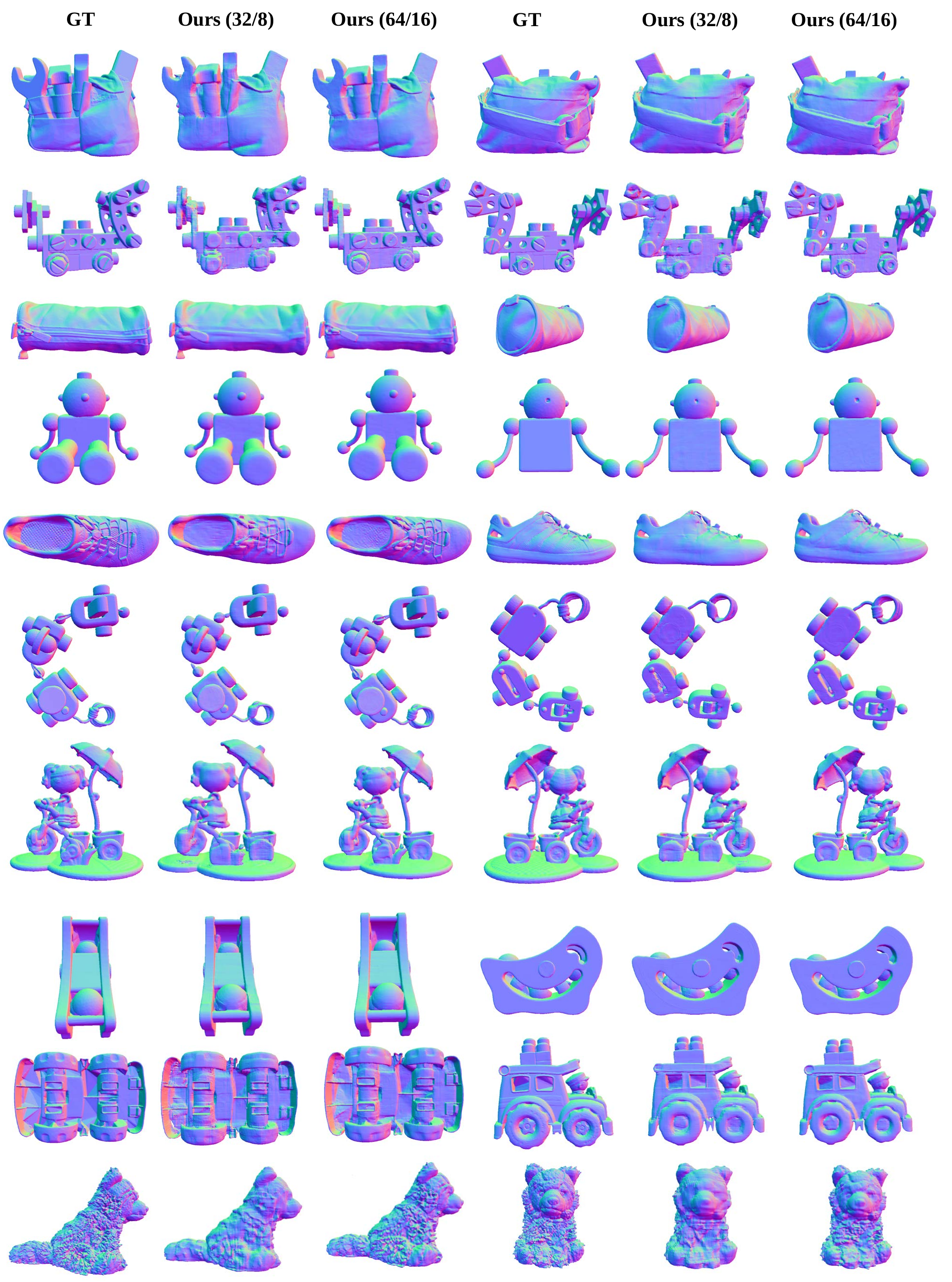}
    \vspace{-2mm}
    \caption{More results of our model.}
    \vspace{-4mm}
    \label{fig:more2}
\end{figure*}

We present more comprehensive comparisons in \cref{fig:more_vis_appendix}, including the reconstruction of more complex objects. Additionally, we compare our method with 3DShape2VecSet~\cite{shape2vecset}, Direct3D~\cite{wu2024direct3d}, and Trellis~\cite{xiang2024structured}. As demonstrated in the examples of the shoe rack, table lamp, and basket, our model exhibits superior capability in reconstructing high-frequency geometric details compared to existing state-of-the-art 3D shape VAEs. For more visualizations of our models, please refer to \cref{fig:more2}.

\section{Limitations and Future Work}
Despite the strong performance of Hyper3D in enhancing 3D shape reconstruction, our approach still has several limitations that present opportunities for future research.

\textbf{Lack of Powerful Generative Models.}  Although Hyper3D provides a robust latent space for 3D representation, we primarily evaluate it in the context of shape reconstruction and integrate it with a naive DiT model. A key future direction is to explore its integration with more powerful generative models for 3D generation tasks.

\textbf{Potential for Multi-Modal Extensions.} Currently, Hyper3D is designed for shape-based encoding and does not explicitly incorporate texture or material information. Many real-world applications require both geometry and appearance modeling to generate fully realistic 3D content. A promising direction for future work is to extend Hyper3D to multi-modal 3D representations by integrating latent texture maps or radiance fields into the hybrid latent structure. This could enable high-quality, photorealistic 3D asset generation beyond pure geometric reconstruction.

While Hyper3D makes significant strides in enhancing 3D shape encoding and reconstruction, it remains an intermediate step toward fully realizing high-fidelity 3D generative modeling. Future research directions include expanding its generative capabilities, improving efficiency, handling more complex geometries, and integrating multi-modal features. These extensions would further unlock Hyper3D’s potential in broader 3D applications, from content creation to real-time 3D synthesis.



\end{document}